\documentclass{article}

\usepackage{arxiv}
\usepackage{xcolor}
\usepackage[utf8]{inputenc} 
\usepackage[T1]{fontenc}    
\usepackage{hyperref}       
\usepackage{url}            
\usepackage{booktabs}       
\usepackage{amsfonts}       
\usepackage{nicefrac}       
\usepackage{microtype}      
\usepackage{lipsum}
\usepackage{graphicx}
\graphicspath{ {./images/} }

\title{Captioning Visualizations with Large Language Models (CVLLM): A Tutorial}

\author{
 Giuseppe Carenini \\
  Professor and Director of the Master in Data Science, Department of Computer Science\\
  University of British Columbia\\
  V6T 1Z4, Vancouver, BC, Canada\\
  \texttt{carenini@cs.ubc.ca} \\
   \And
 Jordon Johnson \\
  Lecturer, Department of Computer Science\\
  University of British Columbia\\
  V6T 1Z4, Vancouver, BC, Canada\\
  \texttt{jordon@cs.ubc.ca} \\
  \And
  Ali Salamatian \\
  Undergraduate Student Research Awards Recipient\\
  University of British Columbia\\
  V6T 1Z4, Vancouver, BC, Canada\\
  \texttt{alisalam@students.cs.ubc.ca} 
}

\begin{document}
\maketitle
\begin{abstract}
Automatically captioning visualizations is not new, but recent advances in large language models (LLMs) open exciting new possibilities. In this tutorial, after providing a brief review of Information Visualization (InfoVis) principles and past work in captioning, we introduce neural models and the transformer architecture used in generic LLMs.  We then discuss their recent applications in InfoVis, with a focus on captioning. Additionally, we explore promising future directions in this field.
\end{abstract}

\section{Introduction}
It is well-established that visualizations have advantages over text-based representations for a number of analysis tasks, since they more fully leverage our innate visual processing capabilities. However, it has also been found that visualizations can be well-supported by textual augmentations such as captions \cite{stokes2022textoftenbetterthemes}.  Further, recent advances 
in large language models (LLMs) have resulted in their incorporation into an unprecedented number of applications and domains. That being the case, this tutorial aims to provide: (1) an overview of captioning visualizations and key concepts in Information Visualization (InfoVis), (2) an introduction to neural networks and transformers, (3) an exploration of the limitations of LLMs and recent developments in the field, and (4) the latest research on InfoVis captioning using LLMs and Large Vision-Language Models (LVLMs).

We will begin with an overview of key concepts in InfoVis and captioning visualizations, including marks, channels, and content characterization. Following this, we will delve into the underlying mechanisms of LLMs, specifically neural networks and transformers. Finally, we will connect these concepts to discuss recent advancements in visualization captioning using LLMs and LVLMs, exploring their limitations and highlighting promising future research directions.

\section{Past Editions, Similar Initiatives and Target Audience}
This tutorial was first presented at AVI 2024.

A related tutorial, "NLP+Vis: NLP Meets Visualization," was offered at EMNLP 2023 and InfoVis 2022, covering a broader scope including InfoVis for NLP model interpretability and text analytics \cite{joty-etal-2023-nlp}. Our tutorial narrows the focus to the background knowledge and the latest methods used in textual support for InfoVis.

The target audience is researchers and practitioners in visual interfaces who want to understand the fundamental concepts and techniques involved in LLM-based textual support for visualizations.

\section{Organization and Duration}
\label{sec:headings}
The tutorial is presented in two parts, each lasting 90 minutes with a 30 minutes break in between.
\subsection{Part 1}
The main goal of this part is to lay the necessary background knowledge required to understanding InfoVis and LLMs.
\subsubsection{Key InfoVis Concepts: Abstractions, Marks, Channels}
In visualization design, it is desirable to identify and work with data abstractions and intended user tasks \cite{munzner2014visualization}. This section explores the importance of task abstraction and methods to achieve it using marks (geometric primitives) and channels that control the appearance of these marks.

\begin{figure}
  \centering
  \includegraphics[scale=0.3]{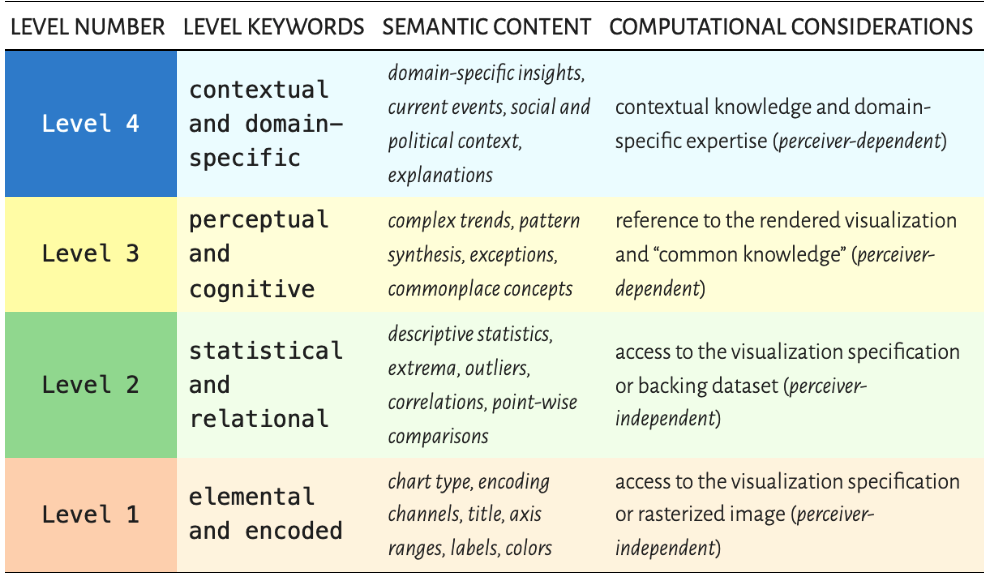}
  \caption{"A four-level model of semantic content for accessible visualization. Levels are defined by the semantic content conveyed by natural language descriptions of visualizations." \cite{2022-vis-text-model}}
  \label{fig:fig1}
\end{figure}

\begin{figure}
  \centering
  \includegraphics[scale=0.5]{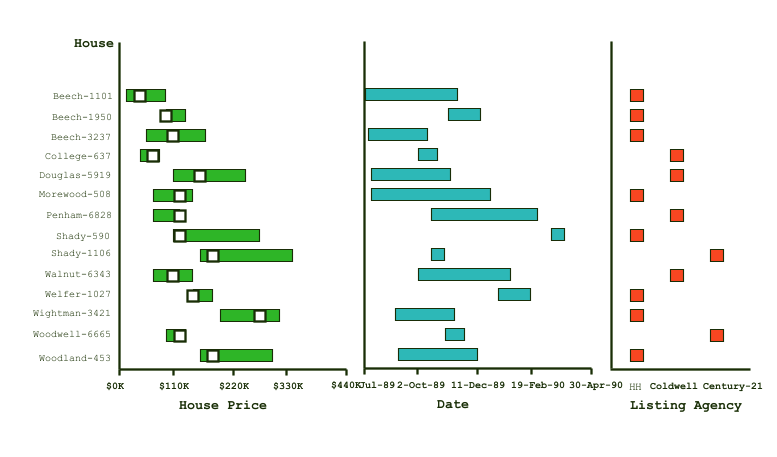}
  \caption{"The Y-axis identifies the houses in the three charts. In the left chart, house prices are shown along the X-axis. The house’s selling price is shown by the left edge of the bar, whereas the house’s asking price is shown by the right edge of the bar..." \cite{mittal-etal-1998-describing}.}
  \label{fig:fig2}
\end{figure}

\subsubsection{Captioning visualizations }
Visualizations with numerous attributes are often difficult to understand completely until explained. Captions provide interpretations of visualizations that help readers understand the purpose of a visualization. They can improve recall and comprehension of depicted data and are crucial for individuals with visual impairments. Moreover, text augmentation of charts can be used in search and question answering. 

In this tutorial, we explain the four levels of semantics used in captions (as depicted in Figure \ref{fig:fig1}), and examine the extent in which each level is covered from early visualization captioning work (e.g., shown in Figure \ref{fig:fig2}) to more recent advancements such as LSTM encoder-decoder models, transformers, and the generated captions using language models (e.g., shown in Figure \ref{fig:fig3}).

\subsubsection{Neural Networks and the Transformer architecture}
In this section, a wide range of fundamental concepts are explained. Firstly, the neural networks are introduced in the context of next token prediction \cite{poole2023artificial}. We then cover methods to improve simple neural models, such as the attention mechanism \cite{jurafsky2023speech}, and discuss the transformer architecture \cite{guardian2023chatbots} used in generic LLMs \cite{minaee2024large}. 

\subsection{Part 2}
The main goal of this part is to explain the limitations of LLMs, explore methods to mitigate these limitations, and discuss the latest developments in visualization captioning.

\subsubsection{Large Language Models: Limitations and Recent Development \cite{minaee2024large}}
In this section, we show that although LLMs have become amazingly proficient at language competence, they are not nearly as good at functional competence  such as solving arithmetic and novel planning problems, often involving issues like hallucinations \cite{openai2024gpt4} that can negatively impact InfoVis captioning. LLMs, essentially being very large neural networks also suffer form lack of interpretability.

We conclude this section by discussing some of the latest techniques developed to address the above limitations, such as Chain-of-Thought (CoT) \cite{wei2023chainofthought, kojima2023large, liévin2023large}, Retrieval-Augmented Generation (RAG) \cite{gao2024retrievalaugmented}, and Reinforcement Learning from Human Feedback (RLHF) \cite{ouyang2022training, manning2023emnlp}. Lastly, we touch upon advancements in LVLMs \cite{li2022blip} and multimodal models \cite{wu2023nextgpt, geminiteam2024gemini}, which can be effectively applied to InfoVis captioning. 

\subsubsection{Recent Advances and Challenges in InfoVis Captioning: A Review of Key Papers}
In this section, we review six recent papers, five of which were carefully selected from Huang et al.'s survey and GitHub page listing key works in the field \cite{huang2024pixels}. These papers illustrate significant advancements in InfoVis Captioning. Our review covers essential steps in recent research progress, including the creation of novel large datasets and the development and testing of new techniques designed to enhance the quality of generated captions, many of which were discussed in the previous section.

\begin{figure}
  \centering
  \includegraphics[scale=0.3]{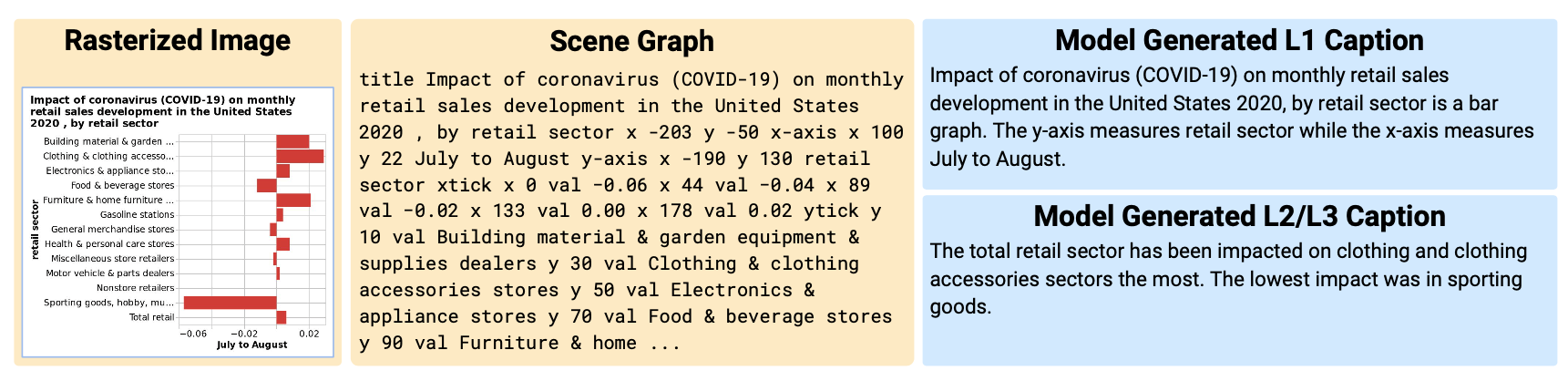}
  \caption{"The scene-graph model’s output L1 caption and L2/L3 caption for a VisText bar chart..." \cite{tang2023vistext}}
  \label{fig:fig3}
\end{figure}

\begin{figure}
  \centering
  \includegraphics[scale=0.75]{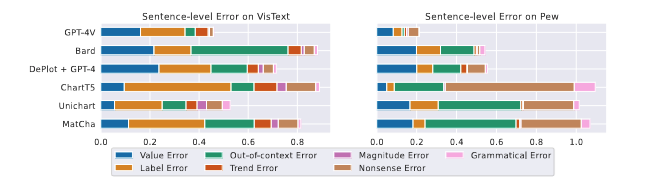}
  \caption{"Error distribution for different models on VisText and Pew." \cite{huang2024lvlms}}
  \label{fig:fig4}
\end{figure}

We begin by discussing the significant contributions of Kantharaj et al. (2022) \cite{kantharaj2022charttotext}, who introduced a large dataset of 44,096 items, including charts, data tables, and captions (primarily at level 2 of semantic content described in Figure \ref{fig:fig1}) along with a benchmark for chart captioning. 
Next, we examine Tang et al.'s (2023) \cite{tang2023vistext} contribution, which includes a dataset of 12,441 items, comprising charts, scene-graphs, data tables, and structured captions. Notably, this dataset incorporates levels 2 and 3 (see Figure \ref{fig:fig1}) through crowdsourcing. As a result, their proposed models fine-tuned on this dataset could generate semantically rich captions as shown in Figure \ref{fig:fig3}.
After that, we highlight the two tasks introduced by Li et al. (2024) \cite{li2024multimodal}: multiple figure and contextualized captioning. Subsequently, we explore how previously mentioned techniques, such as Chain-of-Thought (CoT) and context retrieval have enabled Liu et al.(2024) \cite{liu2024chartthinker} to perform step-by-step learning more effectively and answer relevant questions more accurately. Moreover, we show that through a novel RLHF method, Singh et al. (2023) \cite{singh2023figcapshf} have optimized a generative figure-to-caption model for reader preferences. Finally, we present Huang et al. (2024) \cite{huang2024lvlms} comprehensive typology of factual errors and their finding that state-of-the-art language-vision models, including GPT-4V, frequently produce captions containing with factual inaccuracies as demonstrated in Figure \ref{fig:fig4}. 

In the concluding part of this section (which also concludes the tutorial), we discuss open issues in InfoVis captioning. These include:
\begin{enumerate}
    \item Handling domain specific visualization s (e.g., pathway flowcharts in chemistry) and more complex visualizations (e.g., involving both spatial and temporal features)
    \item Developing more robust and comprehensive evaluation metrics
    \item Enhancing the interpretability of captioning models
    \item Advancing multilingual chart captioning capabilities
\end{enumerate}

\section{Presenters' Past Experiences}
Giuseppe Carenini has taught dozens of undergraduate and graduate courses in his academic career in CS, AI and NLP. Specifically on tutorials in conferences, he has created and given the following ones: \noindent
\begin{itemize}
    \item NLP for Conversations: Sentiment, Summarization, and Group Dynamics; Developed in collaboration with G. Murray and S. Joty -- given by G. Murray at COLING 2018, Santa Fe, New Mexico, August 20, 2018  \cite{murray-etal-2018-nlp}
    \item Discourse Processing and Its Applications in Text Mining; Developed in collaboration with G. Murray and S. Joty \cite{joty-etal-2019-discourse}
    \begin{itemize}
    \item given by G. Carenini and S. Joty at ICDM 2018, Singapore, November 18, 2018
    \item given by G. Carenini and S. Joty at ACL 2019, July 28, 2019
    \end{itemize}
\end{itemize}
 Jordon Johnson is a Lecturer in the UBC CS department, and he has been teaching there since 2018. He has received a teaching award from the CS department and repeated recognition letters from the UBC Faculty of Science. As a graduate student, he received a Killam Graduate Teaching Assistant award, which is the highest such award granted at UBC. While he has taught a variety of courses spanning all undergraduate levels, his most frequently taught courses include CPSC 322 (Introduction to Artificial Intelligence) and CPSC 422 (Intelligent Systems).

\bibliographystyle{unsrt}  
\bibliography{references}  
\end{document}